\documentclass[10pt,twocolumn,letterpaper]{article}

\usepackage{ijcb}
\usepackage{times}
\usepackage{epsfig}
\usepackage{graphicx}
\usepackage{amsmath}
\usepackage{amssymb}

\usepackage{helvet}
\usepackage{courier}
\usepackage{multirow}
\usepackage{setspace}
\usepackage{subfigure}
\usepackage{url}
\usepackage[boxed,commentsnumbered]{algorithm2e}
\usepackage{algorithmicx}
\usepackage{algpseudocode}
\usepackage{dsfont}
\usepackage{sidecap}

\usepackage{array}

\newcolumntype{V}{>{$\vcenter\bgroup\hbox\bgroup}c<{\egroup\egroup$}}
\def\Hline{\noalign{\hrule height 4\arrayrulewidth}}

\graphicspath{{./imgs/}}

\ijcbfinalcopy % *** Uncomment this line for the final submission

\ifijcbfinal\fi
\begin{document}

%%%%%%%%% TITLE
\title{A Framework for Improving the Performance of Verification Algorithms with a Low False Positive Rate Requirement and Limited Training Data}

\author{Ognjen Arandelovi\'c\\
Pattern Recognition and Data Analytics Group, Deakin University, Australia}

\maketitle
\thispagestyle{empty}

%%%%%%%%% ABSTRACT
\begin{abstract}
In this paper we address the problem of matching patterns in the so-called verification setting in which a novel, query pattern is verified against a single training pattern: the decision sought is whether the two match (\textit{i.e.}\ belong to the same class) or not. Unlike previous work which has universally focused on the development of more discriminative distance functions between patterns, here we consider the equally important and pervasive task of selecting a distance threshold which fits a particular operational requirement -- specifically, the target false positive rate (FPR). First, we argue on theoretical grounds that a data-driven approach is inherently ill-conditioned when the desired FPR is low, because by the very nature of the challenge only a small portion of training data affects or is affected by the desired threshold. This leads us to propose a general, statistical model-based method instead. Our approach is based on the interpretation of an inter-pattern distance as implicitly defining a pattern embedding which approximately distributes patterns according to an isotropic multi-variate normal distribution in some space. This interpretation is then used to show that the distribution of training inter-pattern distances is the non-central $\chi^2$ distribution, differently parameterized for each class. Thus, to make the class-specific threshold choice we propose a novel analysis-by-synthesis iterative algorithm which estimates the three free parameters of the model (for each class) using task-specific constraints. The validity of the premises of our work and the effectiveness of the proposed method are demonstrated by applying the method to the task of set-based face verification on a large database of pseudo-random head motion videos.
\end{abstract}

%%%%%%%%% BODY TEXT
\section{Introduction}\label{s:intro}
\begin{figure*}[Htb]
  \centering
  \subfigure[Retrieval]{\includegraphics[height=0.1\textwidth]{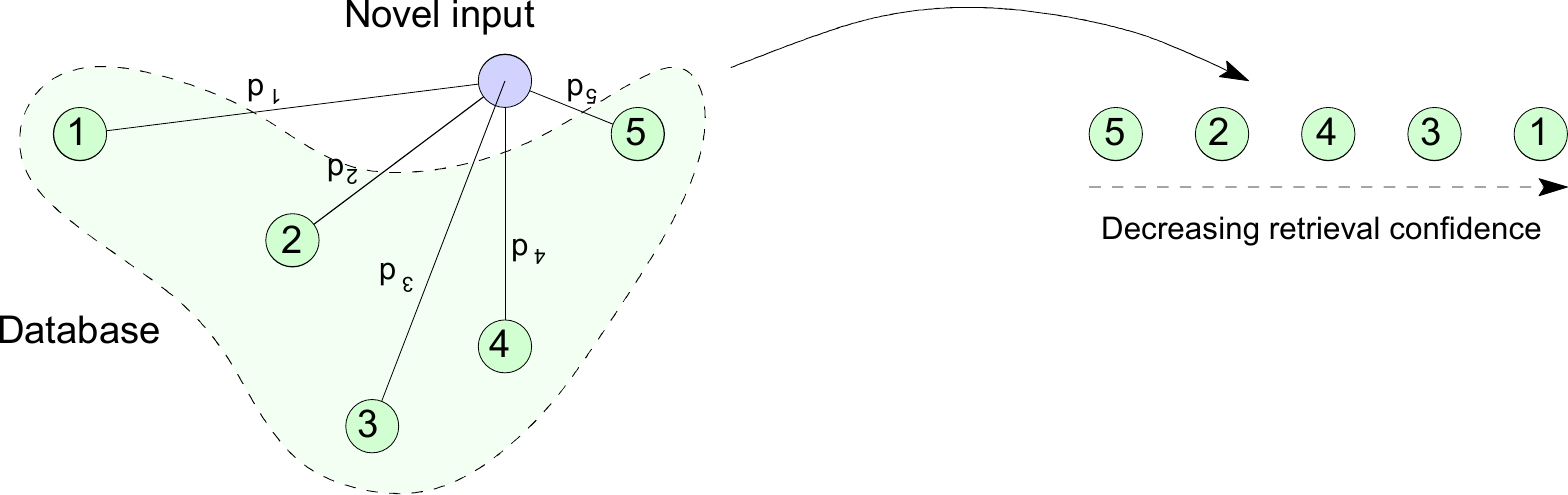}}~~~~~~~~~~~~
  \subfigure[1-to-N]{\includegraphics[height=0.1\textwidth]{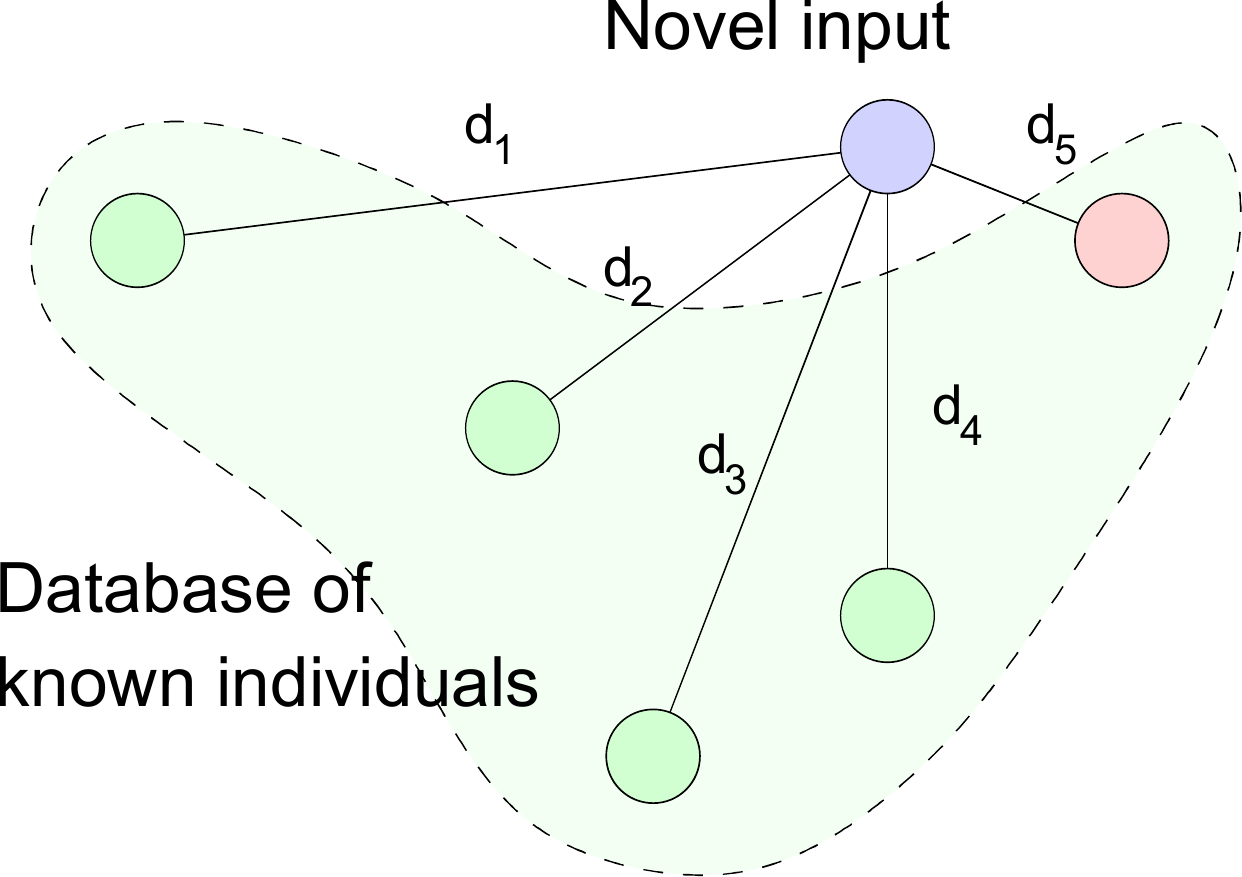}}~~~~~~~~~~~~
  \subfigure[1-to-1 (verification)]{\includegraphics[height=0.1\textwidth]{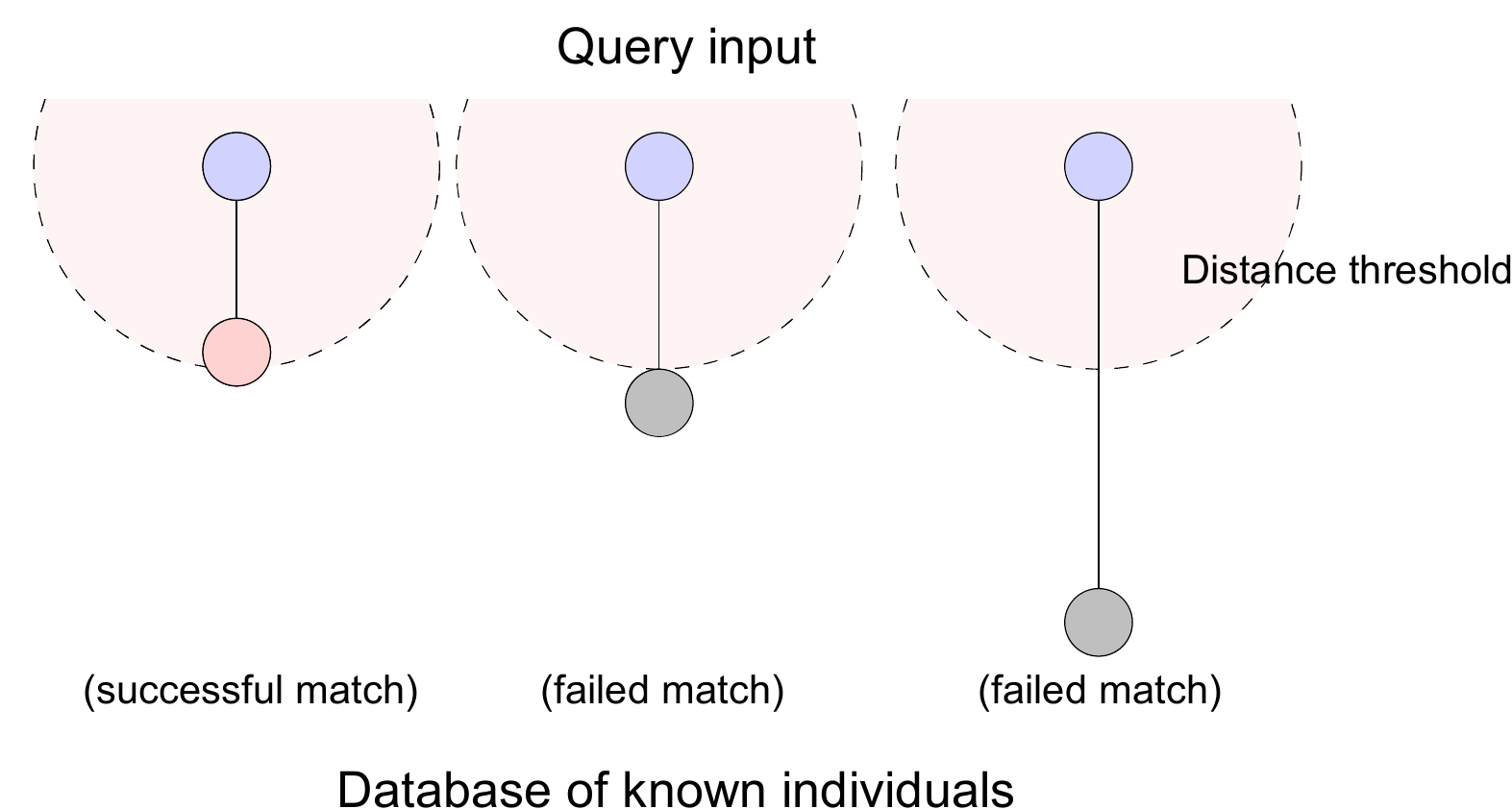}}
  \vspace{10pt}
  \caption{Conceptual illustration of retrieval, 1-to-N, and 1-to-1 (or verification) matching paradigms. }
  \label{f:paradigms}
\end{figure*}

In computer vision terminology, the category of 'recognition problems' is understood to describe a variety of matching paradigms. The retrieval paradigm, for example, refers to the comparison and \emph{ordering} of data instances or patterns (such as images, faces, videos) from a database according to their similarity to a given query pattern~\cite{AranZiss2011,PhilChumIsarSivi+2007}; this is illustrated conceptually in Figure~\ref{f:paradigms}(a). In contrast, in the 1-to-$N$ recognition setting, the query pattern is compared against the patterns of $N$ known classes and matched with the one with the highest similarity~\cite{Flyn2008,Pari2011}, as seen in Figure~\ref{f:paradigms}(b). The present work concerns itself with the verification paradigm whereby the query is compared against a single database pattern and a decision made on whether or not the two patterns belong to the same class~\cite{CaoYinTangSun2010,TolbElBaElHa2006}, as in Figure~\ref{f:paradigms}(c). This problem setting is often encountered in the practical deployment of biometric systems -- a user wishing to access a particular resource, such as a building or a computer, (i) declares his identity as one of the authorized users, (ii) provides biometric information used to verify this claim, and (iii) is granted or denied access.

\subsection{Verification paradigm for matching} The central component of all of the mentioned recognition scenarios can be understood as comprising an algorithm for determining the similarity (or, equivalently, a pseudo-distance) between two patterns, and a decision structure surrounding it. Thus, in verification, after the similarity between the query pattern and the target database pattern is computed, it is compared with a specific threshold. The patterns are successfully matched if and only if the computed similarity exceeds the threshold.

Across a broad spectrum of applications, verification related research to date has virtually exclusively focused on the first stage in the pipeline, \textit{i.e.}\ the development of more sophisticated distance measures between patterns. In the field of face recognition, for example, there has been and continues to be a vast research effort devoted to increasing the robustness of matching to the most pervasive confounding factors encountered in practice, such as illumination~\cite{TanTrig2010}, pose~\cite{ZhanGao2009}, and facial expression changes~\cite{AlOsBennMian2009}, as well as partial occlusion~\cite{WrigYangGaneSast+2009}. As this example illustrates, finding a robust and discriminative distance measure is in most cases a major challenge. In contrast, the subsequent thresholding process appears straightforward. Indeed, in principle this is the case -- all that is needed is for the performance characteristics of the chosen similarity measure to be established empirically to allow the choice of a threshold suitable for a particular application.

\subsection{Verification threshold selection} 
\begin{figure}[thp]
  \centering
  \includegraphics[width=0.22\textwidth]{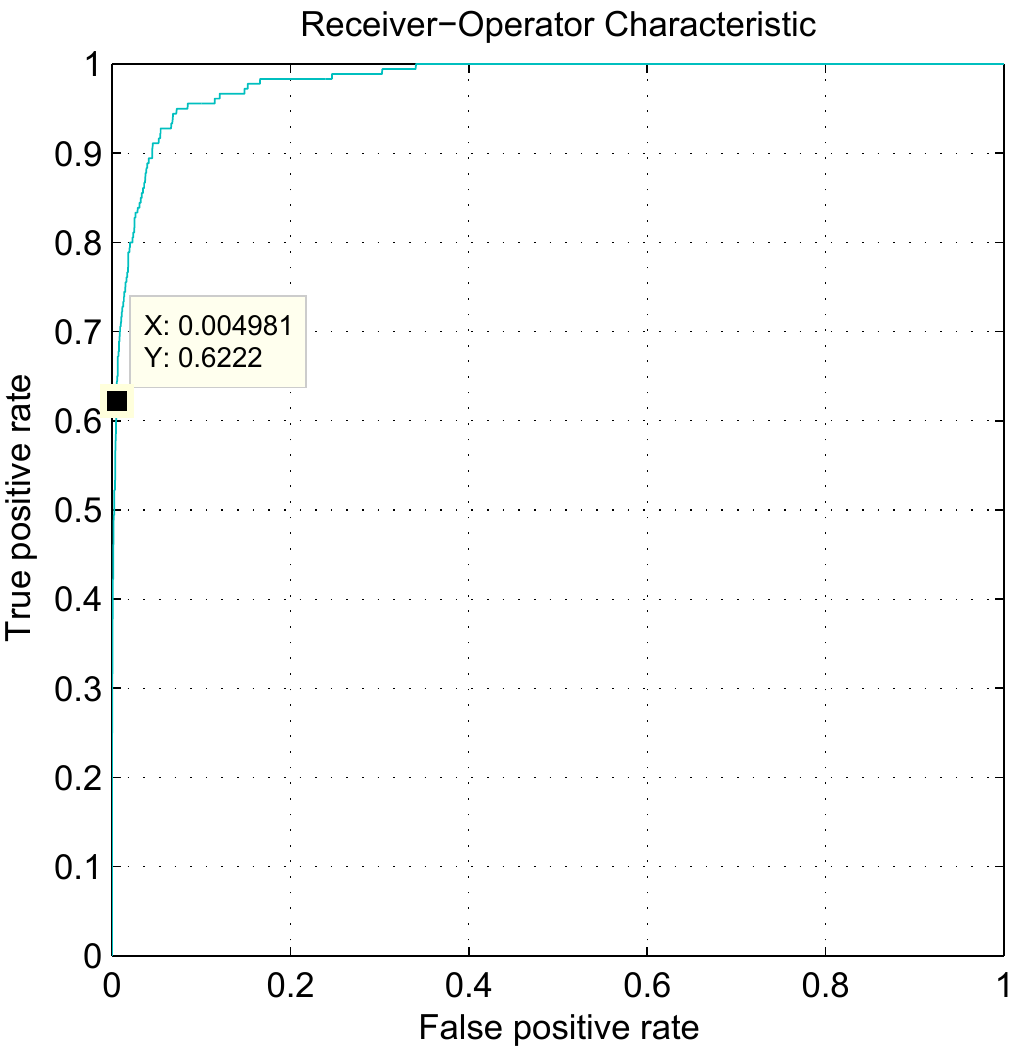}
  \vspace{10pt}
  \caption{The receiver-operator characteristic (ROC) curve, showing the relationship between the true positive (acceptance) rate and the false positive rate. The characteristic was established by evaluating a commercial face recognition algorithm FaceIt~\cite{HeoAbidPaik+2003} on a database of loosely constrained head motion video sequences (see Section~\ref{s:eval}) -- approximately 5,000 pattern comparisons were used (\textit{i.e.}\ similarities between sets of detected faces). The point highlighted on the curve corresponds to a `high security' operating mode for which the false positive rate is relatively low (approximately 0.005 or 0.5\%, which means that 5 in 1000 intruders are admitted). }
  \label{f:roc}
\end{figure}

However, in practice the task of selecting a threshold which conforms to specific operational requirements is in fact far from trivial. The primary reason for this is to be found in the limited amount of data available when a verification algorithm is deployed and trained under realistic conditions. This constraint most significantly affects those cases in which a verification algorithm needs to operate in a `high security setting' \textit{i.e.}\ achieve a low false positive rate (FPR), as illustrated in Figure~\ref{f:roc}. It is not difficult to see why this is the case -- from the very nature of the requirement it follows that the vast majority of inter-class comparisons on the training data set will be far from the desired threshold and as such contribute little information towards its correct inference. Thus, the contribution of the present work builds on the premise that additional knowledge and thus increased robustness of the estimate of the desired threshold can be achieved by the use of a statistical model of inter-class distances. The specific form of the model adopted here is that of a non-central $\chi^2$ distribution. In the next section we explain this choice in detail, including the assumptions which underlie it, and propose an automatic method for inferring its parameters from limited training data available to a verification algorithm. The performance of our approach is examined on the problem of set-based face recognition on a large data set of head motion sequences in Section~\ref{s:eval}.

\section{Statistical model based on the non-central $\chi^2$ distribution}\label{s:intro}
Let us begin by describing the specific problem that this paper addresses. We consider a verification system and assume that there are $N$ classes known to it. A pattern corresponding to each class is used to train the system. To use a familiar biometric example, each of the known classes may be a person who has been enrolled and should be allowed access by the system, and the corresponding training pattern may be an image of the person's face, a motion sequence, a voice recording, or a fingerprint. A novel pattern with an associated claim of the membership to one of the known classes is then used to query the system in an attempt to be granted access. This decision is made by a verification algorithm which computes a measure of similarity between the novel pattern and the training pattern of the class in question.

\subsection{Threshold specificity}\label{ss:specificity} Firstly, let us observe that the similarity threshold used need not be universal across the database and in fact, ideally it \emph{should not} be. Rather, the threshold should be class-specific. For example, if a particular person has markedly uncommon appearance, the same rate of false positive errors can be achieved using a higher similarity threshold than in the case of somebody with a common, closer to ``average'' appearance~\cite{BrucBurtDenc1994}. Although this phenomenon is well known and widely researched in the experimental psychology and neuroscience communities, it is seldom exploited in computer based verification systems. The likely reason for this can be found in the central problem examined in this paper and described in the previous section -- if the verification threshold is set on a class-specific basis, the amount of available information for its inference is reduced even further. For example, if a purely data driven approach is adopted on the data set used to produce the plot in Figure~\ref{f:roc}, there is hardly any information which can be utilized directly. This observation reinforces our motivation for a statistical model driven method instead. A small but notable corpus of work on this problem includes approaches which utilize ideas from the extreme value theory, such as that of Scheirer \textit{et al.}~\cite{ScheRochMichBoul2011} (also see work on multi-modal fusion by Aggarwal \textit{et al.}~\cite{AggaRathBollChel2008} and Poh \textit{et al.}~\cite{PohMeraKitt2009}). Although related this work does not stand as a direct alternative to the method described in our manuscript. There are several reasons for this. Firstly these methods consider a correct match to a probe to be an outlier in the generic, non-match distribution. This is not a reasonable premise in many applications including face recognition. Both from theory and experimentally we can say that the face space is smoothly populated; indeed, Figure~3 illustrates this well. This directly contradicts the key assumption of Scheirer \textit{et al.}  We make no similar assumption. Secondly, the existing methods merely consider the best matching (by some distance criterion) image and rejects it as the correct match if it is not an outlier of the generic, non-match distribution. For this reason they are not capable of doing what we address in this paper: the problem of choosing a particular distance threshold to produce a specific FAR. Indeed this is witnessed by the nature of the evaluation reported by Scheirer \textit{et al.}~\cite{ScheRochMichBoul2011}.

\subsection{The proposed model}\label{ss:proposed} Our idea is to interpret the distances produced by the verification algorithm as corresponding to an embedding of patterns in some high-dimensional space (note that we assume that the distances are in the range $[0,\infty)$ -- in the case of measures which are instead confined to a finite interval, such as $[0,1]$, this can be achieved using a simple transformation by a \textit{logit}-like function~\cite{Cram2003}).  Then our model assumes that the distribution of possible class patterns of all classes is associated with a random variable $\mathbf{X}$ with the corresponding probability density function in the form of an isotropic multi-variate normal distribution; a random sample drawn from $\mathbf{X}$, and a sample set corresponding to a single class, are illustrated conceptually in three dimensions in Figure~3. A smaller point cloud (in blue) is also shown on the same plot -- these are samples from a single class only. Changing the verification matching threshold for this class can be viewed as varying the diameter of the hypersphere centred at the class mean and used to separate the class of interest from the remaining classes.

\begin{figure}[htb]
  \centering
  \includegraphics[width=0.22\textwidth]{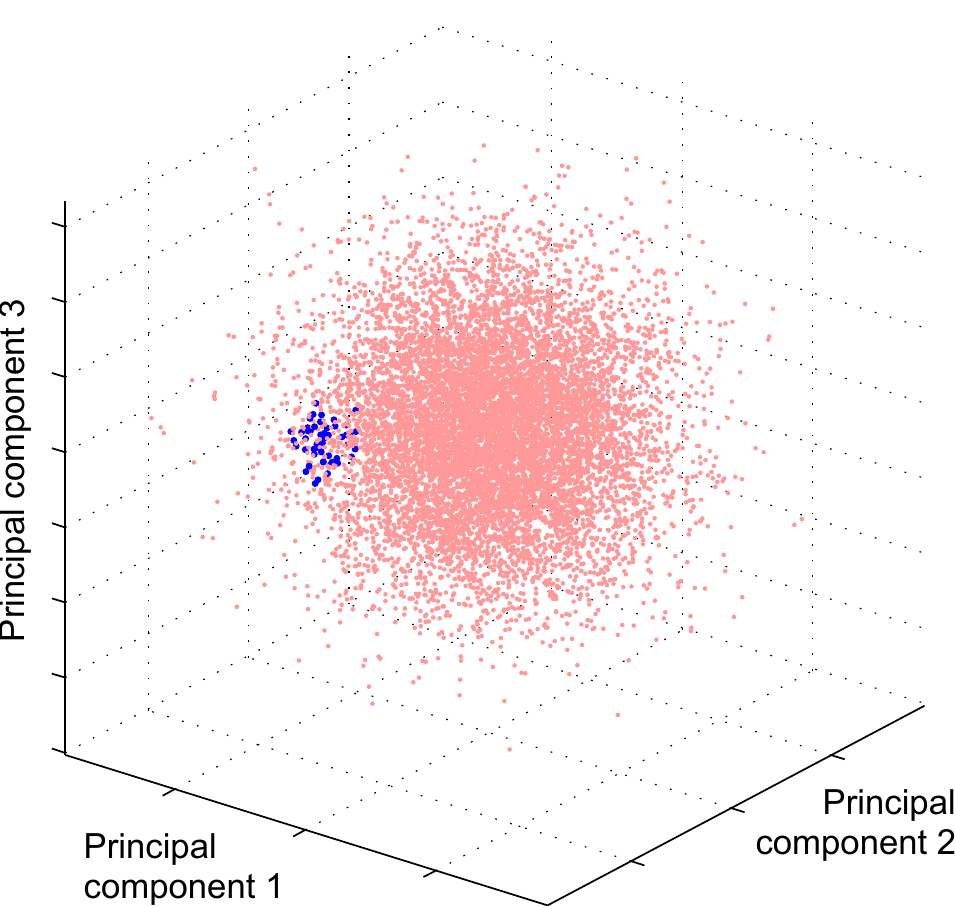}\hspace{20pt}
  \label{f:gauss}
  \vspace{15pt}
  \caption{A conceptual illustration of the initial premise of the present work -- that the distance measure used to quantify the similarity of different patterns in a verification system can be understood as corresponding to the Euclidean distance in a space and embedding in which the patterns are distributed according to an isotropic multivariate normal distribution. Red points represent samples across classes drawn from this distribution, while blue points correspond to samples of one class only which is itself also underlain by an isotropic multivariate normal distribution.}
\end{figure}

Without loss of generality, let us choose the training pattern of the first class, $c_1$, as the origin of the embedding space. Using the proposed model, the distance between $c_1$ and the training pattern of the $i$-th class can be written as:
\begin{align}
  {\delta_i}^2 = \sum_{j=1}^{\text{dim}} \left( {\pi_i}_j \right)^2
  \label{e:dist1}
\end{align}
where ${\pi_i}_j$ is the component of $x_i$ in the direction of the $j$-th orthonormal axis. We can also write:
\begin{align}
  \left( \frac{\delta_i}{\sigma} \right)^2 = \sum_{j=1}^{\text{dim}} \left( \frac{{\pi_i}_j}{\sigma} \right)^2
  \label{e:dist2}
\end{align}
where $\sigma$ is the normalizing factor such that $\sigma \mathbf{X}$ is underlain by the standard isotropic multi-variate normal distribution. Under the assumption that the distribution of instances of each class in this space is also governed by an isotropic multivariate normal distribution, each ${\pi_i}_j$ itself is also normally distributed and the right-hand side of Equation~\ref{e:dist2} can be readily recognized as corresponding to a non-central $\chi^2$ distribution (which we will denote by the symbol ${\chi'}^2$) \cite{Haga1999}:
\begin{align}
  \left( \frac{\delta_i}{\sigma} \right)^2 \sim {\chi'}^2(\text{dim},\lambda).
  \label{e:dist3}
\end{align}

Recall that the non-central $\chi^2$ distribution is defined by two parameters, the number of degrees of freedom $k$ and the non-centrality parameter $\lambda$, with the corresponding probability density function which can be written as:
\begin{align}
  p_{{\chi'}^2}(x;k,\lambda)=\frac{1}{2}~e^{-(x+\lambda)/2} \left(\frac{x}{\lambda}\right)^{k/4-1/2} I_{k/2-1}(\sqrt{\lambda x}),
\end{align}
where $I_a(x)$ is a modified Bessel function of the first kind:
\begin{align}
  I_a(x)=\left(\frac{x}{2}\right)^a~\sum_{i=0}^\infty \frac{ (x^2/4)^i } {i!~\Gamma(a+i+1)}.
\end{align}

The result in Equation~\ref{e:dist3} suggests a straightforward approach to determining the matching threshold for verification of a particular class which satisfies a constraint on the false positive rate:
\begin{enumerate}
  \item compute the distances between the training pattern of the class of interest and the training patterns of all other classes,
  \item determine the parameterization of the non-central $\chi^2$ distribution which best explains the computed distances,
  \item select the distance threshold as that for which the value of the cumulative density function corresponding to the distribution from the previous step is equal to the desired false positive rate.
\end{enumerate}

\subsection{Model parameter estimation}\label{ss:fitting}
Although conceptually simple, the approach for determining the class-specific verification matching threshold outlined in the previous section is challenging in practice. The key source of difficulty is to be found in the problem of model parameter inference from data. In our case additional difficulty is posed by the unknown spatial scaling parameter $\sigma$ -- notice that we observe values of $\delta_i$ in Equation~\eqref{e:dist1} and not the ${\chi'}^2$ distributed $\delta_i/\sigma$ in Equation~\eqref{e:dist2}.

To summarize the task, we wish to infer the values of $\sigma$ and $\lambda$, as well as $\text{dim}$, from the observed distances $\{\delta_1,\ldots\}$. Considering that this cannot be achieved in the closed form, we adopt an iterative approach instead. To make the problem tractable, we recognize and utilize the practicability of imposing several constraints.

Firstly, we assume that from prior understanding of a specific task, bounds (upper and lower) can be placed on the parameters $\sigma$ and $\text{dim}$. Indeed, in practice their values can usually be restricted to quite a narrow range. For example, in Section~\ref{s:eval} from prior work on the analysis of human faces we adopt the constraint that the dimensionality of the face space is 15--22~\cite{Lewi2004}, \textit{i.e.}\ $15 \leq \text{dim} \leq 22$. By virtue of this, and considering that different combinations of possible values of $\sigma$ and $\text{dim}$ are explicitly tested against the available data, the fitting of the non-central $\chi^2$ distribution is reduced to the inference of a single parameter: the centrality parameter $\lambda$. This is a common problem encountered across different fields of research, from signal processing to social sciences. Although no closed form solution for the optimal value exists, a number of different approximations have been described in the literature~\cite{GorrHeimHodgGree2006,Lope2000,SaxeAlam1982}. In the present paper we adopt the recently proposed approximation of Li \textit{et al.}~\cite{LiZhanDai2009}. In summary, the estimate $\lambda^*$ of the non-centrality parameter $\lambda$ given a set of samples $\{\delta_1,\ldots\}$ is computed as:
\begin{align}
  \lambda^*(\{\delta_1,\ldots\}) = max\left\{ \bar{\delta} - \text{dim}, \beta \bar{\sigma} \right\}
\end{align}
where $\bar{\delta}$ is the data mean, $\beta$ a free parameter of the estimator given by (as discussed by Li \textit{et al.}~\cite{LiZhanDai2009}):
\begin{align}
  \beta = \frac{1}{1+\text{dim}},
\end{align}
and $\text{dim}$, as before, the number of degrees of freedom of the corresponding $\chi^2$ distribution.

\begin{figure}[thb]
  \centering
  \subfigure[Best fit, optimal parameters ($\rho=0.9996$)]{~~~~~~\includegraphics[width=0.35\textwidth]{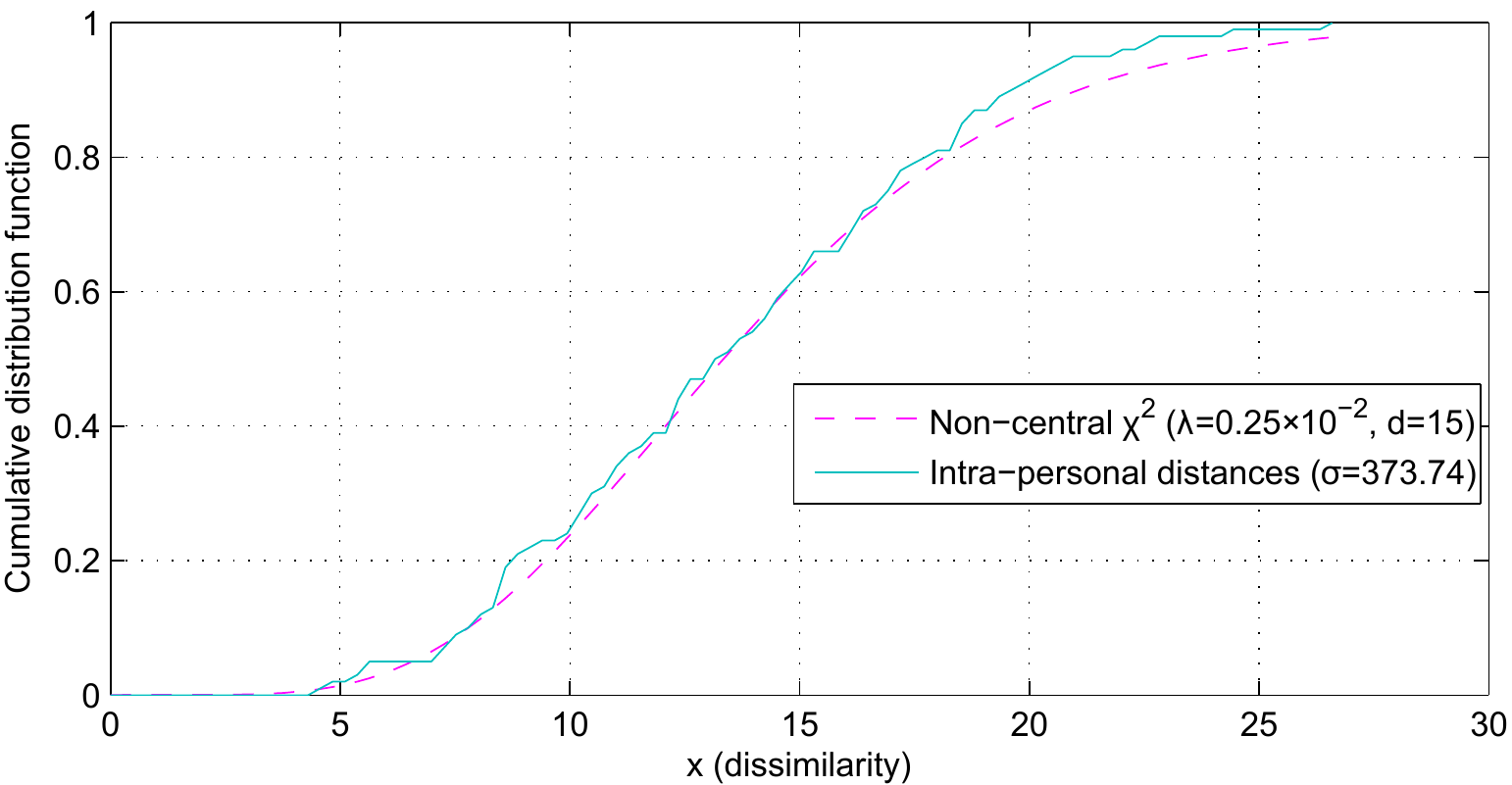}\label{f:fit1}~~~~~~}
  \subfigure[Best fit, sub-optimal parameters ($\rho=0.9261$)]{~~~~~~\includegraphics[width=0.35\textwidth]{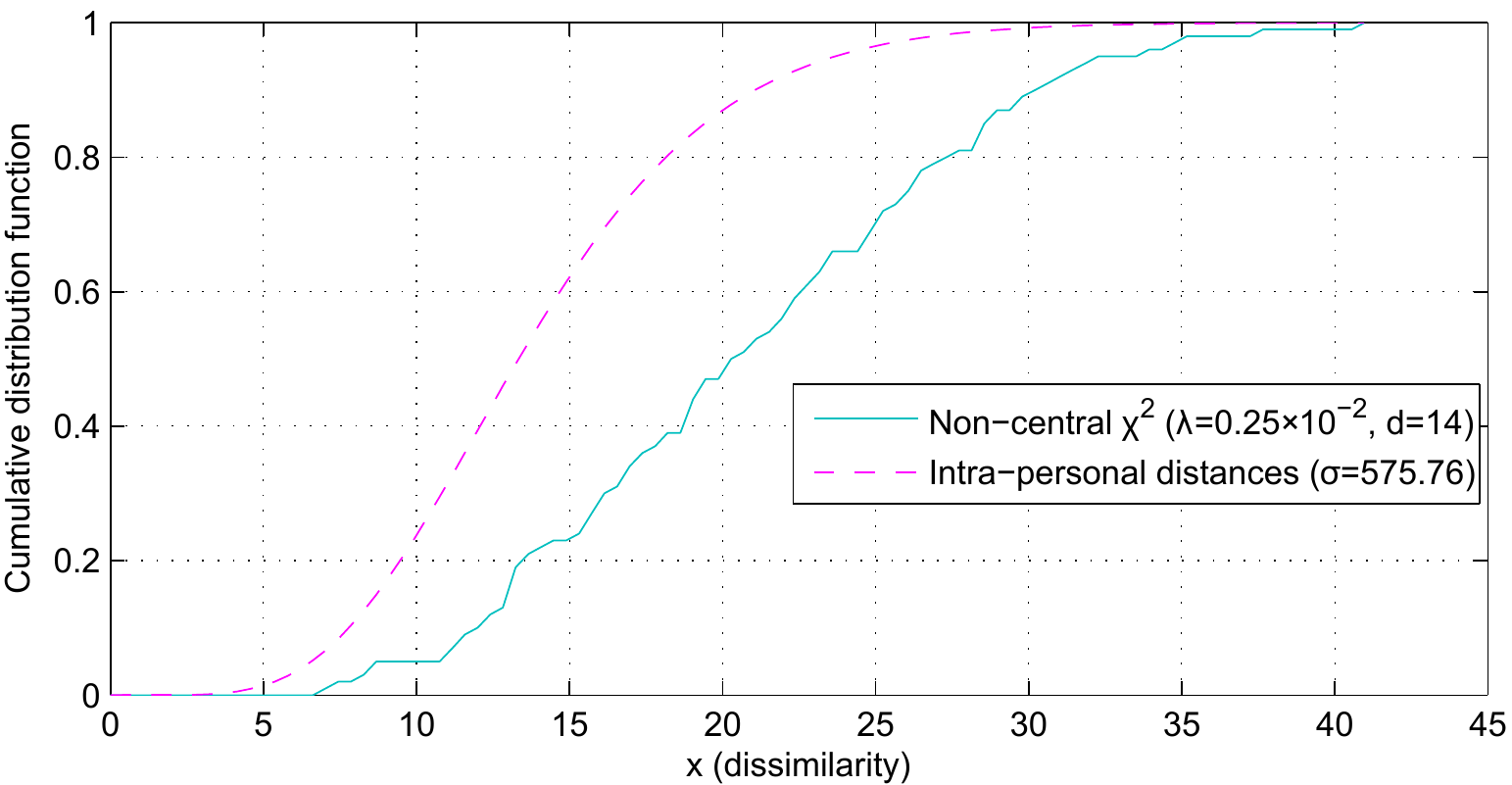}\label{f:fit2}~~~~~~}
  \vspace{10pt}
  \caption{ Plots of the of the empirical cumulative density function estimate on an example data set (solid cyan lines) compared with non-central $\chi^2$ distribution cumulative density function (dashed purple lines) using (a) the all round best-fit set of parameter values, and (b) an intermediate result, \textit{i.e.}\ the best fit for the non-centrality parameter $\lambda^*$ (computed in the inner-most loop in Algorithm~\ref{a:fitting}), and a non-optimal combination of values for the spatial scaling factor $\sigma$ and the number of free parameters $\text{dim}$ (two outer loops in Algorithm~\ref{a:fitting}). }
  \vspace{-10pt}
\end{figure}

Lastly, we assess the quality of a particular fit using the Bhattacharyya distance between the estimates of the cumulative distribution function, one being a direct empirical estimate and the other explicitly given by a particular set of parameters. More specifically, we evaluate the Bhattacharyya distance using samples at all $\delta_i/\sigma$ at which the empirical density estimate becomes simply:
\begin{align}
  \phi_{\text{ref}}(\delta_i/\sigma) = \left| \left\{ (\delta_j/\sigma) ~|~ \delta_j \leq \delta_i) \right\} \right|
\end{align}
that is, the number of distances in training set smaller than $\delta_i$. The exact, \textit{i.e.}\ the target cumulative distribution function with the parameters $k$ and $\lambda$ is given by:
\begin{align}
  \phi(x;k,\lambda) = e^{-\lambda/2}~\sum_{i=0}^\infty \frac{(\lambda/2)^i}{i!} \phi(x;k,0),
\end{align}
where $\phi(x;k,0)$ is the cumulative distribution function of the (central) $\chi^2$ distribution with $k$ degrees of freedom. Bringing all of the elements of the proposed framework together, the proposed fitting can be concisely written as summarized by Algorithm~\ref{a:fitting}. Figure~\ref{f:fit1} shows the plots of the non-central $\chi^2$ distribution cumulative density function with the best-fit parameters (dashed purple line) and of the empirical cumulative density function estimate on an example data set (solid cyan line). For the sake of comparison, in Figure~\ref{f:fit2} we also plotted the non-central $\chi^2$ distribution cumulative density function with the best-fit non-centrality parameter $\lambda^*$ (inner-most loop in Algorithm~\ref{a:fitting}) for a non-optimal combination of values for the spatial scaling factor $\sigma$ and the number of free parameters $\text{dim}$ (two outer loops in Algorithm~\ref{a:fitting}).

\IncMargin{-0em}
\begin{algorithm}
  \fontsize{10}{16}\selectfont

  \SetAlgoLined
  \KwData{Inter-class distances $\{\delta_1,\delta_2\ldots\}$; constraints on parameters $\dim_\text{low}$ and $\dim_\text{high}$, and $\sigma_\text{low}$ and $\sigma_\text{high}$}
  \KwResult{Parameters $\lambda_\text{opt}$ (non-centrality parameter) and $\dim_\text{opt}$ (number of degrees of freedom), of a non-central $\chi^2$ distribution corresponding to data $\{\delta_1,\delta_2\ldots\}$, and the data scaling parameter $\sigma_\text{opt}$}

  \BlankLine

  $\rho_\text{opt} \gets -1$\;
  \For{$\dim = \dim_\text{low} \to \dim_\text{high}$}{
    \For{$\sigma = \sigma_\text{low} \to \sigma_\text{high}$}{
      $\lambda \gets max\left\{ \bar{\delta} - \dim, \beta \bar{\delta} \right\} \text{ where } \beta = \frac{1}{1+\dim}$\;
      $\phi_{\text{ref}} \gets \text{CumDensity}(\{\sigma \delta_1,\sigma \delta_2\ldots\})$\;
      $\phi \gets \text{CumNonCentralChi2}(\text{dim}, \lambda)$\;
      $\rho \gets \text{NCC}(\phi_{\text{ref}}, \phi)$\;
      \If{$\rho > \rho_\text{opt}$}{
        $\rho_\text{opt} \gets \rho$\;
        $\dim_\text{opt} \gets \dim$\;
        $\sigma_\text{opt} \gets \sigma$\;
        $\lambda_\text{opt} \gets \lambda$\;
      }
    }
  }
  \caption{\small Fit a non-central $\chi^2$ distribution to the set of training inter-class distances.}
  \label{a:fitting}
\end{algorithm}

\section{Evaluation}\label{s:eval}
In this section we report the evaluation results of the methods proposed in the present paper on real-world data. Specifically, we applied the proposed algorithm in the context of face recognition -- a particularly suitable problem given its frequent use of the verification paradigm and a major potential for a wide spectrum of practical applications. For this experiment we used the large data set of face motion video sequences in different illuminations described in~\cite{Aran2012}. There are two key reasons why we chose this particular data set. Firstly, it has been repeatedly demonstrated that the scope of extrinsic variability contained within it presents a major challenge both to state of the art algorithms in the literature, as well as commercial software~\cite{AranCipo2013}. Secondly, by adopting a set matching paradigm (an increasingly active research area in its own right) we were able to use a baseline distance measure very different than the Euclidean distance, thus removing any doubt over the generality of our results and the key premises of our work that they support (please refer back to Section~\ref{ss:proposed} for detail).

\subsection{Baseline setup}
The University of Cambridge face data set contains face motion video sequences of 100 individuals (varying ages and ethnicities). For each person in the database there are 7 sequences of the person performing loosely constrained, pseudo-random motion (signiﬁcant translation, yaw and pitch, negligible roll) for 10~s, acquired at 10~fps. Each sequence was acquired in a different illumination setting.

For each person, a single training (`enrollment') sequence was used as a training pattern (as described in Section~\ref{s:intro}). These training sequences were used for the learning described in Section~\ref{ss:fitting} \textit{i.e.}\ for determining class-specific (person-specific) verification thresholds. All subsequent querying of the algorithm was performed in the 1-to-1 fashion \textit{i.e.}\ a single novel set was compared with a single training set, and the two either matched as belonging to the same person or to different people. To assess the performance, we adopt the following evaluation paradigm. At the time of training, the desired (target) false positive rate is specified, allowing for the class-specific thresholds to be determined as explained in Section~\ref{ss:proposed}. These thresholds are then used with the remainder of the data, withheld from training. Each pattern (face set) is used in a verification test against every class and the false positive rate determined by counting how many verification attempts which do not correspond to the same person are erroneously admitted by an algorithm.

For the between-set distance measure we adopt the canonical correlations based approach (sometimes also referred to as the mutual subspace method) previously used by a number of authors \textit{e.g.}~\cite{FukuStenYama2006} (also see~\cite{BachJord2005,Aran2014}). In summary, this approach consists of: (i) the application of principal component analysis to each image set, (ii) the representation of a set by the projection matrix $\mathbf{P}_i$ comprising the first $d_p$ principal directions (we used $d_p=6$), and (iii) the computation of the distance between two sets as $\delta = 1-\max_{\mathbf{u},\mathbf{v}} \left\{ (\mathbf{P}_i \mathbf{u})^T (\mathbf{P}_j \mathbf{v}) \right\}$, which can be performed efficiently using singular value decomposition~\cite{BjorGolu1973}. As stated earlier, we sought to adopt a distance measure as different in nature than the Euclidean distance. Our aim was to demonstrate the universality of the assumptions underlying the proposed model, and in particular the interpretation of $\delta$ as implicitly defining an embedding under which it becomes Euclidean, as detailed in Section~\ref{ss:proposed}.

Lastly, we compared the proposed method with two data-driven alternative approaches for estimating the verification thresholds, one generic (the same threshold is used for all classes, as explained in Section~\ref{ss:specificity}) and the other class-specific. For both methods the thresholds are determined by linear interpolation. Specifically, if $N$ inter-class distances are used in training, the threshold corresponding to the false positive rate $\epsilon$ becomes:
\begin{align}
  \tau_{\epsilon} =
    \begin{cases}
       \frac{\epsilon}{1/N}~\delta_{i_{\epsilon+}}   & \text{ for } \epsilon \leq 1/N\\
       \delta_{i_{\epsilon-}} + \frac {N \epsilon - \lfloor \epsilon N \rfloor} {\lceil \epsilon N \rceil-\lfloor \epsilon N \rfloor}~(\delta_{i_{\epsilon+}} - \delta_{i_{\epsilon-}})
                    & \text{ for } \epsilon   >  1/N
    \end{cases}
\end{align}
where $\delta_{i_{\epsilon-}}$ and $\delta_{i_{\epsilon+}}$ are respectively the $\lfloor \epsilon N \rfloor$-th and $\lceil \epsilon N \rceil$-th lowest training distances.

\subsection{Results}
The key results of our experiments are summarized in Table~\ref{t:res}. As the table shows, we evaluated the proposed method and compared it with the two baseline approaches while varying the target false positive rate from 0.5\% (\textit{i.e.}\ 1 intruder admitted per 200 intrusion attempts) down to 0.05\% (1 intruder admitted per 2000 intrusion attempts). For each method and a target false positive rate, the table shows the actual false positive rate attained by the method on unseen data (on the left-hand side of the corresponding column), and the ratio of the target and actual intrusions allowed (on the right-hand side of the corresponding column).

\begin{table*}
  \centering
  \renewcommand{\arraystretch}{1.5}
  \caption{ Summary of evaluation results. A row in the table corresponds to an instance of our experiment with the corresponding target false positive rate (FPR) specified in the leading column. For each of the three methods compared we show the actual FPR attained when data unseen in training was used to query the verification system (on the left in the corresponding column), and the ratio of the actual and target FPRs (on the right in the corresponding column). }
  \vspace{8pt}
  \begin{tabular}{c||cc|cc|cc}
    \Hline
     \multirow{3}{*}{Target FPR}            & \multicolumn{6}{c}{Thresholding method}\\
    \cline{2-7}
       & \multicolumn{2}{c|}{Generic data-driven} & \multicolumn{2}{c|}{Class-specific data-driven} & \multicolumn{2}{c}{Proposed}\\
    \cline{2-7}
               & FPR     & Ratio                & FPR     & Ratio                   & FPR     & Ratio\\
    \hline
      0.50\%      & 1.52\%  & 3.0                  & 0.02\%  & 0.05                    & 0.91\%  & 1.8 \\
      0.25\%     & 0.79\%  & 3.1                  & 0.00\%  & 0.00                    & 0.39\%  & 1.6 \\
      0.10\%      & 0.36\%  & 3.6                  & 0.00\%  & 0.00                    & 0.13\%  & 1.3 \\
      0.05\%     & 0.19\%  & 3.8                  & 0.00\%  & 0.00                    & 0.06\%  & 1.2 \\
    \Hline
  \end{tabular}
  \label{t:res}
  \vspace{10pt}
\end{table*}

Consistent trends were apparent both within and between methods across different experiments. The generic threshold data-driven approach invariably overestimated the threshold required to meet the target false positive rate, resulting in a significantly greater (3 to 4 times) number of successful intrusions than desired. In contrast, the class-specific data-driven alternative always overestimated the thresholds, resulting in an over-stringent system for the given specification. It is also important to observe that the performance of both methods consistently worsened as the target false positive rate of the system was reduced -- both the overestimation of the threshold by the generic method and the underestimation by the class-specific method became more severe.

In all experiments -- that is, for all values of the target false positive rate -- the proposed method achieved by far the best performance. In the worst case the false positive rate attained by our algorithm was 1.8 times greater than the specified target, while the best performance of the other methods in any of the experiments was a three-fold discrepancy. It is insightful to observe that unlike in the case of the two approaches discussed previously, the performance of our method improved as the value of the target false positive rate was reduced.

\section{Summary and conclusions}
In this paper we considered the problem of matching patterns, each considered to represent a class instance from a set of classes, in the context of a verification framework.
%This means that matching is done on a 1-to-1 basis, that is, by comparing two patterns only, and the decision sought is
Specifically, we focused on the key task of selecting the optimal matching threshold when the operational requirement for the system is one of a low false positive rate. This task was shown to be inherently challenging for data-driven approaches, as by the very nature of the requirement, most of the available training data (in the form of inter-class distances) is too far from the optimal threshold to be informative. Secondly, we argued that in order to optimize verification performance the matching threshold should be set in a class-specific rather than generic manner, and showed that if this approach is adopted the challenge posed by limited training data is increased even further. Thus, a solution in the form of a statistical model of inter-class distance distribution was proposed. The key idea underlying our work was to interpret inter-class distances as implicitly defining a pattern embedding in which the distances become Euclidean. This was shown to give rise to the non-central $\chi^2$ distribution, differently parameterized for each training class. The three free parameters of the model were estimated using a novel algorithm which utilizes task-specific constraints and an analysis-by-synthesis iterative process. Lastly, the validity of the premises of our work, as well as the effectiveness of the proposed methods, was evaluated by applying it to the task of set-based face recognition.

{\small
\bibliographystyle{ieee}
\bibliography{./my_bibliography}
}

\end{document}